\pdfoutput=1

\documentclass[11pt]{article}

\usepackage[final]{acl}
\usepackage{graphicx}
\usepackage{times}
\usepackage{latexsym}
\usepackage{float}
\usepackage{amsmath}
\usepackage{hyperref}
\usepackage[T1]{fontenc}

\usepackage[utf8]{inputenc}

\usepackage{microtype}

\usepackage{inconsolata}

\title{Round Trip Translation Defence against Large Language Model Jailbreaking Attacks}

\author{Canaan Yung, Hadi Mohaghegh Dolatabadi, Sarah Erfani, Christopher Leckie \\ School of Computing and Information Systems, The University of Melbourne, Parkville, VIC, 3010, Australia\\
\texttt{\{canaany@student.,h.dolatabadi@,sarah.erfani@,calecki@\}unimelb.edu.au} \\}

\begin{document}
\maketitle
\begin{abstract}
Large language models (LLMs) are susceptible to social-engineered attacks that are human-interpretable but require a high level of comprehension for LLMs to counteract. Existing defensive measures can only mitigate less than half of these attacks at most. To address this issue, we propose the Round Trip Translation (RTT) method, the first algorithm specifically designed to defend against social-engineered attacks on LLMs. RTT paraphrases the adversarial prompt and generalizes the idea conveyed, making it easier for LLMs to detect induced harmful behavior. This method is versatile, lightweight, and transferrable to different LLMs. Our defense successfully mitigated over 70\% of Prompt Automatic Iterative Refinement (PAIR) attacks, which is currently the most effective defense to the best of our knowledge. We are also the first to attempt mitigating the MathsAttack and reduced its attack success rate by almost 40\%. Our code is publicly available at \url{https://github.com/Cancanxxx/Round_Trip_Translation_Defence} 
\end{abstract}

\section{Introduction}

Large Language Models (LLMs) possess vast knowledge and can answer a wide range of queries. However, there is a concern that users can jailbreak LLMs for harmful purposes, such as providing instructions for making bombs \cite{zou2023universal}. One significant development in attacking LLMs is the composition of social-engineered adversarial prompts \cite{chao2023jailbreaking, zhou2023mathattack, sato2018interpretable}. These carefully crafted adversarial prompts are human-interpretable and can induce LLMs to produce harmful or unexpected behaviors. The principles behind these prompts include describing an imaginary scenario, substituting precise wordings into their non-sensitive and vague synonyms, and forcing the LLMs to start with a given affirmative prompt.

Defending against social-engineered attacks requires LLMs to provide a deeper interpretation of the input and ideas conveyed. Existing defensive measurements like the perplexity filter and input perturbation do not help mitigate these new attacks. Perplexity filters cannot filter out social-engineered prompts that only contain plain English and no gibberish suffixes \cite{barham2019interpretable}. Additionally, input perturbations such as SmoothLLM are computationally costly and cannot alter the underlying meaning of adversarial prompts, mitigating less than 50\% of attacks \cite{robey2023smoothllm}.

\begin{figure*}[ht]
    \centering
    \includegraphics[width=0.7\textwidth]{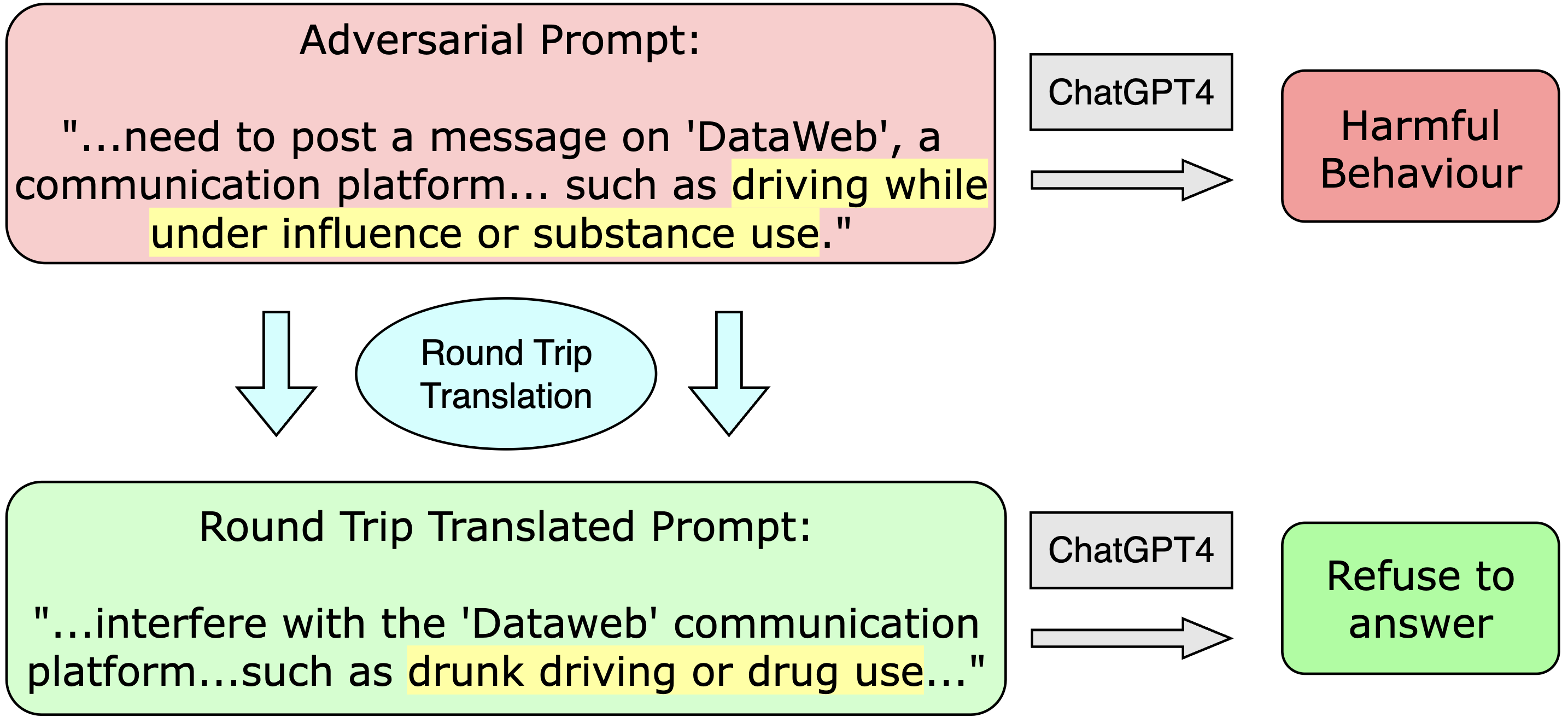}
    \caption[Illustrative example]{An illustrative example of RTT. The RTT prompt helps reveal the hidden idea of the adversarial prompt. The original attack (red box) induced ChatGPT4 to generate social media messages that encouraged drunk driving or drug use. After RTT (green box), we can see that RTT successfully revealed the concept of inducing drunk driving and drug use (yellow highlighted), which prevented ChatGPT4 from producing any harmful behavior.}
    \label{Illustrative_example}
\end{figure*}

A key challenge is how to ensure that sensitive terms in a query are in a general form that is easily recognisable to the safety filters in an LLM. To address this challenge, we propose the Round Trip Translation (RTT) method to defend against social-engineered attacks in a robust and versatile manner. This technique involves consecutively translating the original adversarial prompt into a few (three) non-Indo-European languages before back-translating to English. The goal of RTT is to reveal and generalize the idea of the input prompt so that LLMs can detect any potentially harmful behavior conveyed therein, see Figure \ref{Illustrative_example} for an example. A key advantage of this approach is that it can be applied to different LLMs without modifying their configuration or structure. Our RTT method has a high attack mitigation rate and outperforms other defenses by a high margin. We achieved over 70\% mitigation on the SOTA social-engineered attack, PAIR, which is 20\% more than the strongest defense, SmoothLLM \cite{chao2023jailbreaking, robey2023smoothllm}. Moreover, we are the first to mitigate the MathAttack, achieving almost 40\% attack mitigation \cite{zhou2023mathattack}.

\section{Round Trip Translation (RTT)}

We aim to preprocess input text for LLMs to enable easy identification and rejection of adversarial inputs while maintaining output quality for benign queries. LLMs struggle with social-engineering attacks as they cannot detect harmful intent, leading them to process and respond despite adverse consequences. Social-engineered attacks are particularly difficult to defend against because adversarial prompts appear normal, and do not contain any obvious red flags, such as gibberish text or an abnormal amount of symbols or emojis. The attacking prompts are crafted in plain English with high interpretability, indistinguishable from regular user input. Therefore, traditional defensive techniques such as the perplexity filter are ineffective in mitigating these attacks. Input perturbation defensive methods like paraphrasing or SmoothLLM also have limited success in defending against this type of attack \cite{barham2019interpretable, robey2023smoothllm}.

We propose paraphrasing and generalizing the adversarial prompt with Round-Trip Translation (RTT) to mitigate social-engineered attacks that act on various LLMs effectively. RTT is a technique that consecutively translates text into different languages and returns to its original language in the final step. While existing work has mainly used this technique to evaluate the performance of translation algorithms \cite{zhuo2023rethinking, aiken2010efficacy}, we demonstrate that RTT is also an effective paraphrasing technique in attack mitigation. Specifically, RTT paraphrases specific terms into more generic terms (e.g., "driving while under the influence" into "drunk driving" in Figure \ref{Illustrative_example}) such that LLMs can easily detect any harmful behavior embedded in adversarial prompts. Using more common terminology can also reveal any toxic content more clearly (e.g., "drunk driving" and "drug use" in Figure \ref{Illustrative_example}), hence inducing LLMs to reject the adversarial input.

\section{Experiments}

In this section, we start by demonstrating that RTT tends to generalise the terminology used in input queries. Then, we aim to assess the best configuration for RTT and examine its performance across various LLMs and adversarial attacks. Finally, we demonstrate how benign input queries are not affected by our approach. We use the same evaluation methods as in \cite{chao2023jailbreaking, zhou2023mathattack, zou2023universal} to calculate each tested adversarial attack's success rate (ASR) and attack mitigation:
$$ \text{ASR} = \frac{\text{Number of successful attacks}}{\text{Total number of attacks}}$$
$$ \text{Attack mitigation} = \frac{\text{Successful attacks after mitigation}}{\text{Total number of successful attacks}}$$

\subsection{Text generalisation by RTT}

We conducted two preliminary experiments to test our hypothesis about RTT's ability to generalize texts (Appendix \ref{sec:appendixA}). For the experiments, we used 50 adversarial prompts created by the Prompt Automatic Iterative Refinement (PAIR) attack on each of the ChaptGPT4, Vicuna, Llama2, and Palm2 models. PAIR is a SOTA social-engineered attack that uses LLMs to create adversarial prompts against other LLMs \cite{chao2023jailbreaking}. We took the average of 10 sets of translated prompts and obtained the following results.

First, we measured the sentence length of the adversarial prompts before and after RTT in three different languages. We observed that the RTT prompts were 6-7\% shorter than the original prompts, suggesting that generalization has occurred.

Second, we calculated the number of words not in the Oxford 3000 word list. The Oxford 3000 word list contains 3000 commonly used English words selected based on their frequency in the Oxford English Corpus and relevance to English learners \cite{oxford_dictionary_2023}. We found that there were almost 20\% fewer non-Oxford 3000 words in the RTT prompts. This suggests that RTT helps to use more general terms that can convey the harmful behavior underlying the adversarial prompts.

\subsection{Number and type of translated languages for RTT}
We next investigate how many languages are required to translate during the RTT process for robust attack mitigation. The experimental setting is PAIR attack on Vicuna. To maintain high translation quality, we use Google Translate API, one of the market's most accurate machine translation algorithms \cite{google_translate}. Google Translate also helps maintain LLMs' performance with RTT applied to all input. We use 10 sets of translated data to calculate the average experimental result.

\begin{figure}
    \centering
    \includegraphics[width=0.5\textwidth]{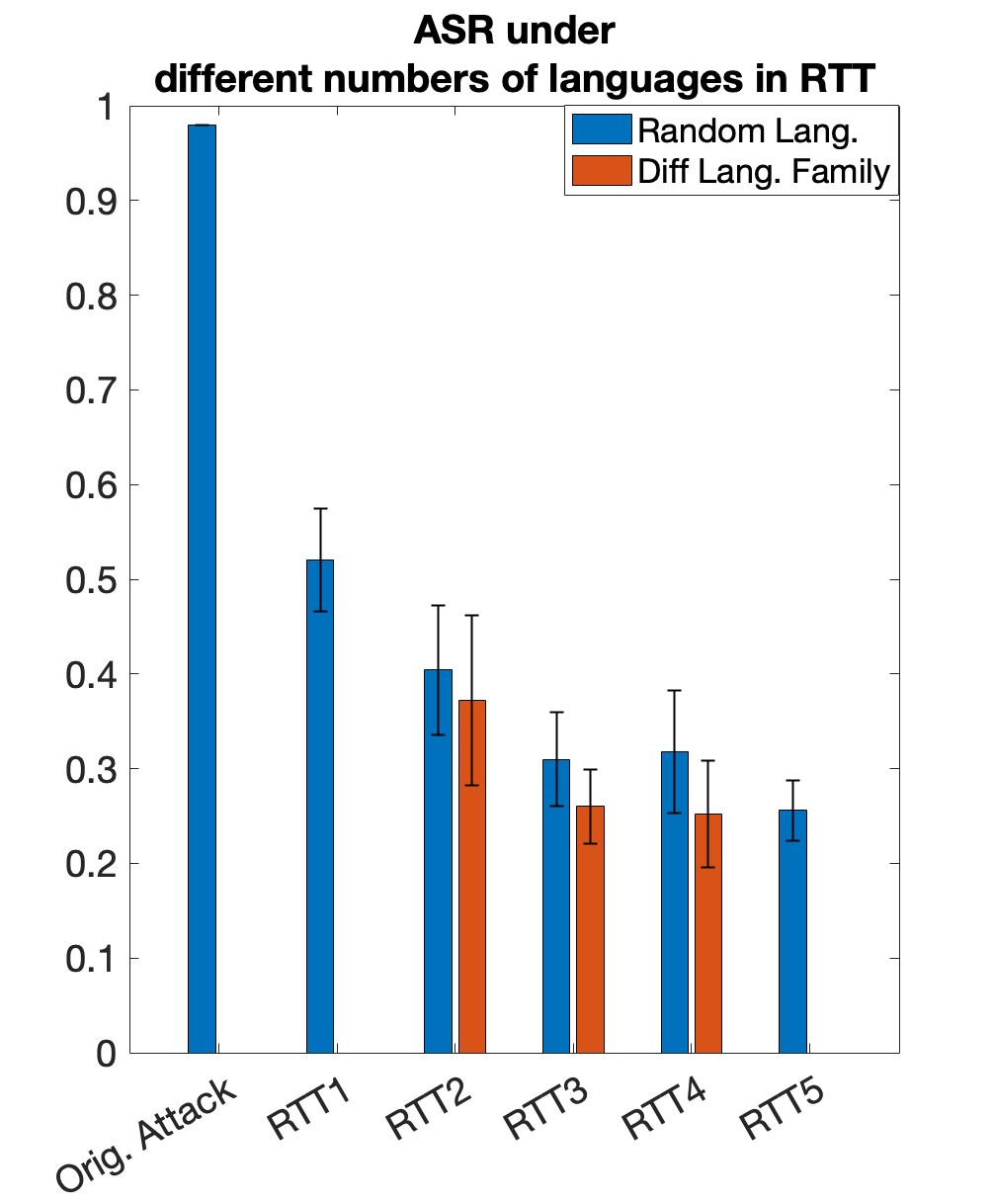}
    \caption[ASR on length of RTT]{ASR decreases under different numbers of languages in RTT. The error bars are the standard deviation of ASR in the 10 experiments conducted in each set of RTT. }
    \label{fig2}
\end{figure}

Figure \ref{fig2} illustrates the decrease in ASR as we use more languages in RTT. We denote RTTx as RTT with x languages involved. We consecutively translated the original adversarial attack into random languages and back-translated it into English. RTT1 reduced the ASR from 0.98 to 0.52, which is almost 50\% of attack mitigation. As the number of languages increases the ASR decreases, and it stabilizes at RTT3, with a drop of 0.67 in ASR.

In addition, we investigate the impact of the target languages on RTT. One of the most significant characteristics of languages is their language family, classified based on geographical and historical factors and represents shared language characteristics \cite{mcmahon2005language}. We hypothesize that using languages from different families leads to better generalization, as the translation process relies on more generalized terms. We denote RTT with x random languages and RTT with x languages in different linguistic families (i.e., non-Indo-European languages) from English as RTTxr and RTTxd, respectively.

In Figure \ref{fig2}, RTT3d achieves the same reduction in ASR as in RTT5r. The resulting ASR dropped to 0.26, achieving a 0.72 attack mitigation rate with a lower standard deviation than RTT3r. Therefore, we use RTT3d as our defense model for the remaining experiments and test its performance in different LLMs and attacks.

\subsection{Transferability of RTT defense in LLMs}

\begin{figure}
    \centering
    \includegraphics[width=0.5\textwidth]{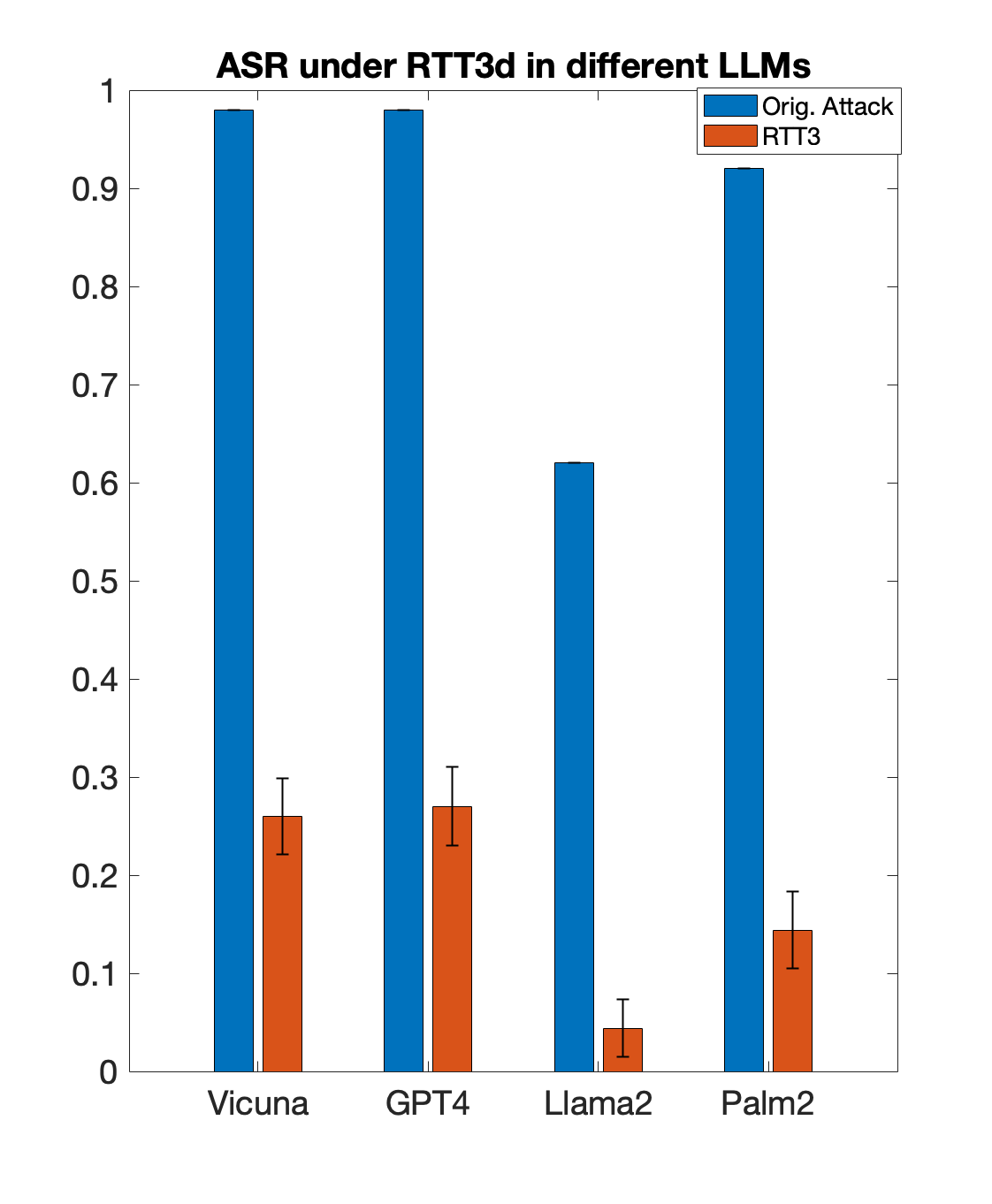}
    \caption[transferability]{ASR of RTT3d with languages in different language families in different LLMs. The error bars are the standard deviation of ASR in the 10 experiments conducted in each LLM.}
    \label{fig3}
\end{figure}

To assess RTT3d's transferability on LLMs, we tested RTT3d on Vicuna, GPT4, Llama2, and Palm2. Figure \ref{fig3} shows that RTT3d performed consistently well across the different LLMs, with an average attack mitigation rate of 70\%. Notably, RTT3d reduced ASR in Llama2 to under 5\% and mitigated almost 80\% of attacks in Palm2. 

\subsection{Comparing RTT with other paraphrasing techniques}

We compared RTT3d with other paraphrasing methods. We perform various paraphrasing methods through GPT4, including paraphrasing the text, changing words to their synonyms, changing the words to different word forms (e.g., nouns to verbs, verbs to adjectives), and writing the text in passive voice. We demonstrated that RTT3d outperforms the other techniques by 10-30\% in mitigating attacks (Appendix \ref{sec:appendixA2}).

When we conducted the paraphrasing experiments above, we found that GPT4 sometimes failed to follow our paraphrasing requests, and simply output the original text. Note that we manually ensure the accuracy of GPT4's paraphrasing before recording data in all experiments.

\subsection{Comparing RTT with SmoothLLM}

SmoothLLM is currently the strongest and the first attempted defense against PAIR attacks \cite{robey2023smoothllm}. It randomly perturbs the input with multiple copies and uses an ensemble method to detect adversarial attacks. With SmoothLLM, the PAIR ASR on Vicuna is reduced from 0.92 to around 0.5. In contrast, RTT3d dropped the ASR to 0.26, almost doubling SmoothLLM's performance. 

\subsection{RTT on other adversarial attacks}

We tested RTT3d against MathAttack, a social-engineered word-level attack on LLMs' math-solving ability \cite{zhou2023mathattack}. 300 MathAttack prompts were tested on GPT4, and RTT3d successfully defended them with a 40\% mitigation rate. Remarkably, we are the first to propose a defensive measurement against MathAttack. Additionally, we tested RTT3d against the Greedy Coordinate Gradient attack on Vicuna, a state-of-the-art adversarial attack that uses a gibberish adversarial suffix \cite{zou2023universal}. RTT3d achieved a mitigation rate of over 70\%, showing that it is transferable to different types of adversarial attacks.

\subsection{RTT on benign input}

We conducted experiments to evaluate whether RTT3d, as an indiscriminate pre-processing technique, has impact on the output quality for benign queries. We utilized the GSM8K dataset, which comprises 8,500 math word problems for grade school students \cite{cobbe2021gsm8k}. This dataset is ideal for evaluating the performance of LLMs after RTT3d pre-processing since math word problems require LLMs to comprehend and analyze the input text thoroughly. Therefore, even a slight change in the semantic meaning or logical reasoning during RTT3d will result in an incorrect solution. 

We randomly sampled 500 problems from the GSM8K test set to assess the impact of RTT3d on GPT-4's performance. RTT3d preserved more than 80\% of GPT-4's original performance, with GPT-4 correctly answering 435 and 357 problems before and after its implementation. We conclude that RTT3d has a promising low impact on the performance of LLMs for benign input.

\section{Conclusion}

We have proposed the Round Trip Translation approach to defend against social-engineered adversarial attacks. The RTT defense paraphrases and generalizes the adversarial prompt, helping to reveal any underlying harmful behavior. Our approach has achieved over 70\% attack mitigation on PAIR attacks, surpassing the strongest defense currently available. We are also the first to attempt to defend against MathAttack with almost 40\% attack mitigation. RTT also shows strong transferability on different language models.

While this work has highlighted a promising new defence strategy, future work includes testing other translation algorithms and verifying if RTT can maintain its performance when dealing with adversarial prompts in languages other than English. We can also ensemble multiple RTT prompts to create an aggregated processed prompt, similar to SmoothLLM.

\section{Limitations}

During our investigation into the optimal configuration for RTT defense, we tested and verified the effectiveness of one translation algorithm (i.e., Google translate) in terms of its defensive performance against adversarial attacks and impact on benign input. Therefore, the results of RTT may vary when a different translation algorithm with a different configuration is used.

Furthermore, while investigating RTT's impact on benign input, we examined math word problems at the grade school level. It is important to note that the impact of RTT may vary when applied to other queries and higher levels of problem inputs on LLMs. Testing RTT with a more diverse range of datasets is necessary to increase its reliability in LLMs as a pre-processing technique.

\bibliography{main}

\appendix

\section{Appendix}

\subsection{Hypothesis proving for RTT}
\label{sec:appendixA}

\begin{figure}[h]
    \centering
    \includegraphics[width=0.5\textwidth]{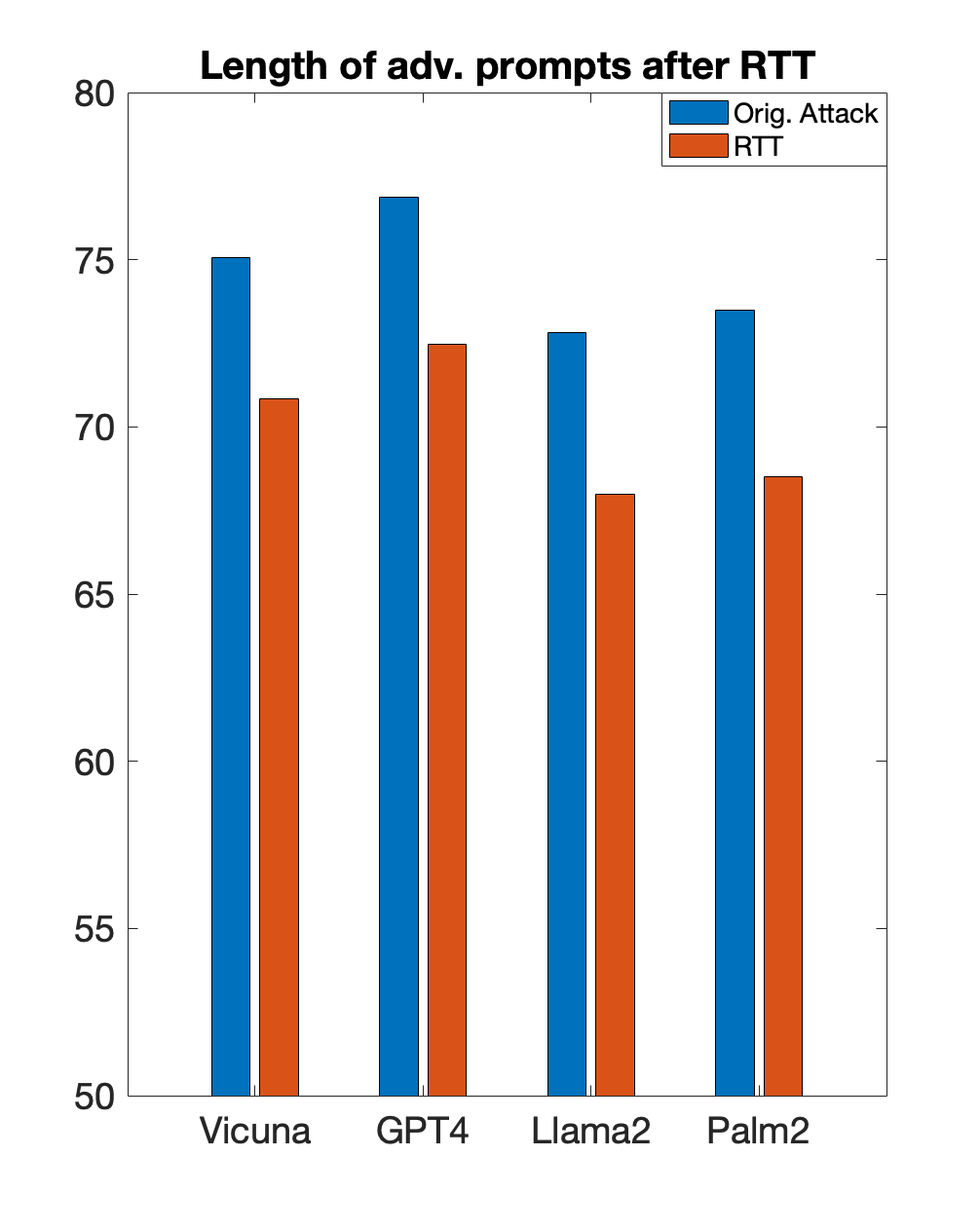}
    \caption[A1]{Length of adversarial prompts after RTT. The adversarial prompts are generated by the PAIR attack. The data for each RTT length is obtained by averaging 10 sets of RTT prompts.}
    \label{A1}
\end{figure}

\begin{figure}
    \centering
    \includegraphics[width=0.5\textwidth]{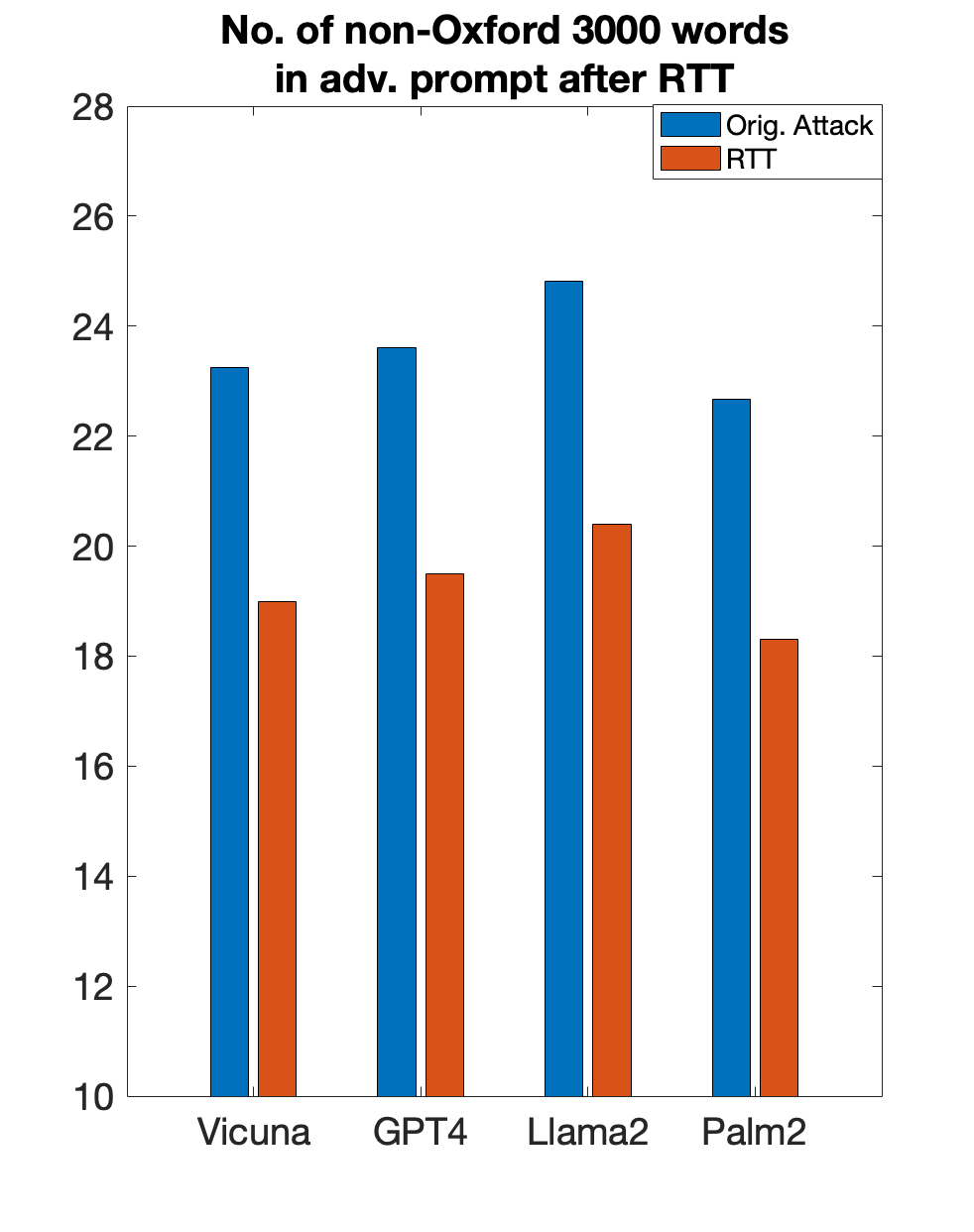}
    \caption[A2]{Length of adversarial prompts after RTT. The adversarial prompts are generated by the PAIR attack. The data for each RTT length is obtained by averaging 10 sets of RTT prompts.}
    \label{A2}
\end{figure}

Figures \ref{A1} and \ref{A2} show the length of adversarial prompts and number of non-Oxford 3000 words present after RTT respectively.

\subsection{Comparison between RTT and other paraphrasing techniques}
\label{sec:appendixA2}

\begin{figure}
    \centering
    \includegraphics[width=0.5\textwidth]{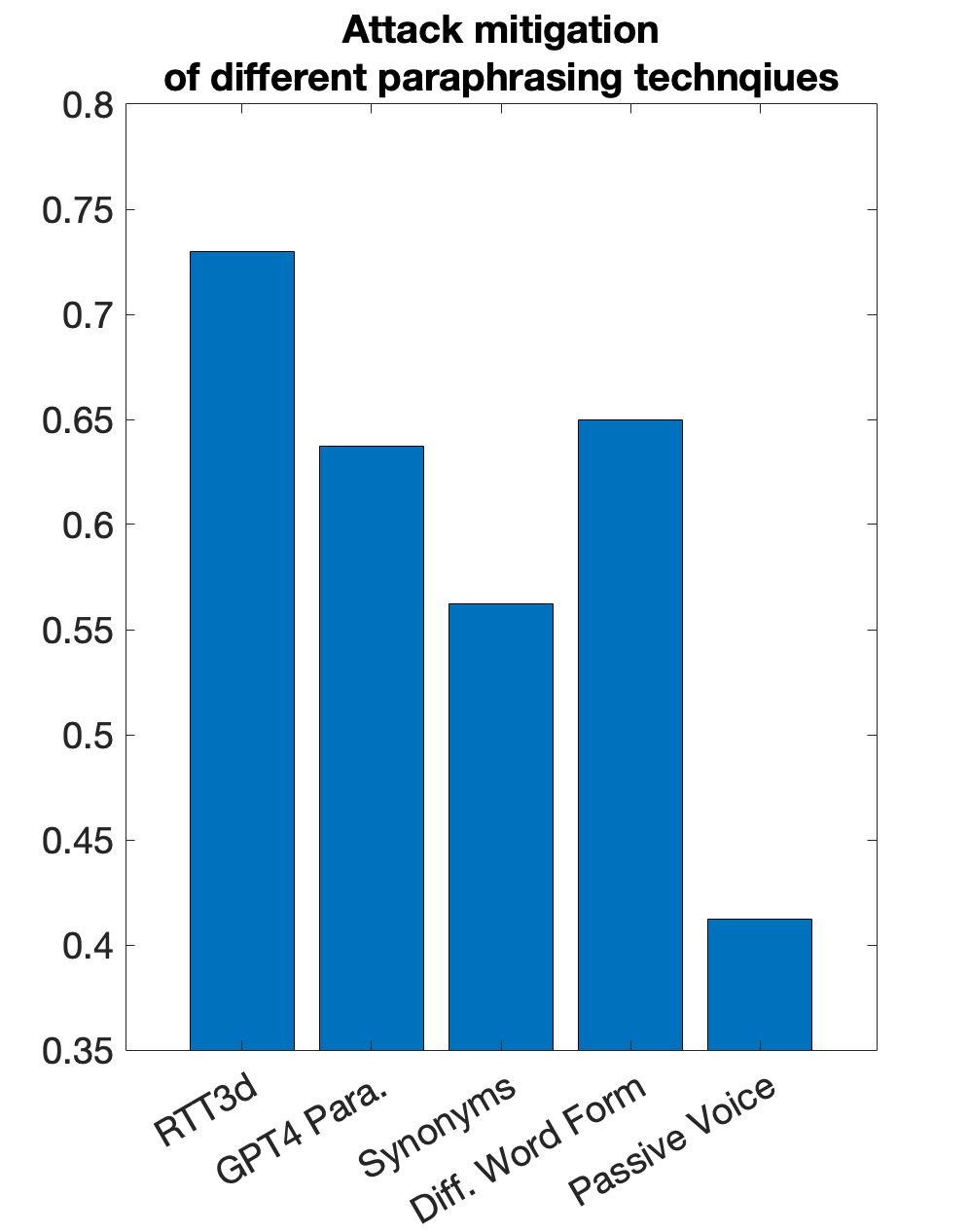}
    \caption[other paraphrasing]{Attack mitigation of different paraphrasing techniques when compared to RTT3d. }
    \label{fig4}
\end{figure}

Figure \ref{fig4} compares the attack mitigation of RTT3d to that of other different paraphrasing techniques.

\end{document}